\documentclass{article}

\usepackage{arxiv}

\usepackage[utf8]{inputenc} % allow utf-8 input
\usepackage[T1]{fontenc}    % use 8-bit T1 fonts
\usepackage{hyperref}       % hyperlinks
\usepackage{url}            % simple URL typesetting
\usepackage{booktabs}       % professional-quality tables
\usepackage{amsfonts}       % blackboard math symbols
\usepackage{amsmath}        % \text{} in math mode
\usepackage{nicefrac}       % compact symbols for 1/2, etc.
\usepackage{microtype}      % microtypography
\usepackage{graphicx}
\usepackage{natbib}
\usepackage{doi}
\usepackage{listings}
\usepackage{xcolor}

\title{Using Large Language Models to Support High-Volume Application Review for an Undergraduate Research Program}

\author{
Varun Aggarwal$^{1,2}$, Kay Kobak$^{1}$, John Howarter$^{1,3}$ \\
$^1$Engineering Undergraduate Research Office, Purdue University \\
$^2$Elmore School of Electrical and Computer Engineering, Purdue University \\
$^3$School of Materials Engineering, Purdue University \\
}

\date{}

\renewcommand{\undertitle}{}
\renewcommand{\shorttitle}{LLMs for High-Volume Application Review}

\hypersetup{
pdftitle={Using Large Language Models to Support High-Volume Application Review for an Undergraduate Research Program},
pdfsubject={cs.CL, cs.CY},
pdfauthor={Author Name},
pdfkeywords={Large Language Models, Application Review, Undergraduate Research, SURF, Automated Essay Scoring},
}

\begin{document}
\maketitle

\begin{abstract}
Undergraduate research programs such as the Summer Undergraduate Research Fellowship (SURF) at Purdue University receive thousands of applications every year, requiring significant time and effort for program staff to evaluate each submission consistently and within tight timelines. This work-in-progress paper describes the development and initial deployment of a large language model (LLM)-based tool to assist in the evaluation of approximately 1,200 student Statements of Purpose (SoPs) for the SURF 2026 cycle at Purdue University. The workflow utilizes OpenAI's GPT models (GPT-4o, GPT-5-mini, and GPT-5.2) and uses a structured rubric across six subcategories, each scored on a 0–3 scale. A few SoPs, graded by program staff, were used to tune the model responses. The model prompt was designed to generate both numerical scores and rationales, both positive and negative aspects, supported by short excerpts from each submission. This paper reports on the workflow and preliminary observations. Using GPT-5.2, the full batch of 1,200 SoPs was processed in approximately 4.6 hours of compute time, averaging roughly 14 seconds per SoP (with per-SoP timing varying with SoP length, which ranged from 500 to 2,000 words). Notable differences in rubric adherence were observed across model versions, with GPT-5.2 adhering most closely. Disagreement in model scores was more pronounced for lower-scoring submissions. The LLM outputs were not used to directly shortlist applicants; rather, they replicated the role previously played by distributed human graders, providing the program coordinator with scored and rationale-annotated outputs for the full applicant pool. The program coordinator then reviewed these outputs alongside each applicant's SoP, applying the same downstream office criteria used in prior SURF cycles, to produce a shortlist of strong candidates. This coordinator review was completed in approximately 4 hours, compared to the multi-week coordination effort required to recruit, train, and reconcile scores from a team of human graders in prior program cycles.

\end{abstract}

% keywords can be removed

\section{Introduction}

Undergraduate research experiences are widely recognized as a critical mechanism for developing STEM talent and improving retention, particularly for underrepresented populations \citep{nasem2017undergrad, linn2015undergrad}. Programs such as the Summer Undergraduate Research Fellowship (SURF) provide structured opportunities for undergraduates to engage in faculty-mentored research. However, as program visibility and institutional enrollment grow, application volumes have increased substantially. For the SURF 2026 cycle at our institution, the program received approximately 3,000 applications, creating a significant operational challenge for program staff.

The traditional review process involves assembling a team of human graders, distributing applications, calibrating reviewers against a shared rubric, collecting scores, and reconciling results. This process is time-intensive not only in the actual reading and scoring of each application, but also in the planning, assignment, and aggregation steps required to coordinate multiple reviewers. Furthermore, when different reviewers bring different levels of expertise and attention, inconsistencies in scoring can arise, a well-documented challenge in peer and panel review processes \citep{recio2022peer}.

Recent advances in large language models (LLMs) have demonstrated their capacity to evaluate written text against structured rubrics with increasing reliability \citep{ramesh2022aes, mizumoto2023exploring, tang2024harnessing}. Studies have shown that with well-designed prompts, rubrics, and calibration examples, LLMs can achieve scoring agreement with human raters that approaches the level of inter-rater agreement among humans themselves \citep{tian2024examining, rodrigues2024assessing}. Critically, research has also shown that LLM evaluation quality is highly sensitive to prompt design, model version, and temperature settings, and that without explicit rubric guidance, LLMs produce inconsistent and unreliable scores \citep{yavuz2025utilizing, garciavarela2025chatgpt}. These findings suggest that while LLMs are not yet a replacement for human judgment, they can serve as a powerful triage and pre-screening tool when deployed thoughtfully.

This work-in-progress paper describes the development of a LLM-based workflow to support the SURF 2026 application review process. The primary objective was to replicate the role of distributed human graders using an LLM, producing scored and rationale-annotated outputs for the full applicant pool that the program coordinator could then review using the same downstream office criteria applied in prior SURF cycles.

\section{Approach}

\subsection{Application Filtering}

Before any LLM-based evaluation, applications were filtered against a set of eligibility criteria established by the program. These criteria included a minimum GPA of 3.2, expressed interest in graduate school, a Statement of Purpose of at least 450 words, ability to attend the full program duration, age of at least 18 years, at least one remaining semester after the program, current standing of at least sophomore, and no prior SURF participation. This initial filtering reduced the applicant pool from approximately 3,000 to approximately 1,200 eligible applications, which then proceeded to SoP evaluation for identifying a pool high potential students.

\subsection{Rubric Design}\label{sec:rubric-design}

The evaluation rubric (Appendix A) was designed specifically for the SURF program and comprised three main categories, Passion, Clarity of Purpose, and Resilience, each containing two sub-categories, for a total of six evaluated dimensions. Each sub-category was scored on a 0-3 scale, yielding a maximum total score of 18. The Passion category assessed Motivation for Scientific Research (whether the applicant articulated what sparked their interest in STEM and why they continue to pursue it) and Initiative (whether the applicant actively sought opportunities beyond mandatory coursework, including opportunities outside their home institution). The Clarity of Purpose category assessed Program Expectations and Benefits (whether the applicant demonstrated a realistic understanding of the SURF program and mentioned specific professional development opportunities, departments, or faculty at Purdue) and Alignment with Future Endeavors and Career Direction (whether the applicant connected the program to future graduate school plans, including specific degree types and fields). The Resilience category assessed Reflection on Learning through Experience (whether the applicant reflected on personal, academic, or research challenges and demonstrated growth) and Problem Solving (whether the applicant described a structured approach to overcoming challenges and sought outside resources). Each score level was accompanied by a detailed behavioral descriptor specifying what that level of performance looks like, providing the model with concrete anchors for its scoring and evaluation rationale.

The current rubric was developed iteratively after observing model outputs during the initial development phase. An earlier version of the rubric, used in prior SURF cycles (Appendix C), relied on more subjective descriptors (for example, ``contagious enthusiasm'' or ``exceptional problem-solving abilities''). We found that the model performed substantially better when the differences between adjacent score levels were defined by objectively observable behaviors rather than qualitative impressions. Replacing subjective anchors with behaviorally specific criteria (for example, ``mentions both their interest in STEM \emph{and} their motivation to continue in STEM,'' or ``applied for opportunities through competitive application processes'') had two effects: the model's scores became grounded across submissions, and the program coordinator was better able to verify each score against the cited evidence in the SoP.

\subsection{Prompt Engineering}

The prompt (Appendix B) instructed the model to evaluate each SoP against the specified rubric and produce a JSON output containing a score and rationale for each of the six sub-categories. The prompt required that each rationale include both positive and negative aspects to justify the assigned score, preventing one-sided evaluations. The prompt also required that rationales be supported by direct excerpts from the SoP. By compelling the model to cite specific applicant excerpts, the program coordinator could quickly verify whether the score was grounded in the actual submission to limit LLM Hallucination.Additionally, the prompt allowed fractional scores for accounting for borderline cases. The complete rubric with all score-level descriptors was embedded directly in each prompt followed by applicant SoP. The SoPs in this study ranged from approximately 500 to 2,000 words in length.

\subsection{Few-Shot Calibration}
To calibrate model behavior against human expectations, approximately three example SoPs at each performance level high ($\text{score}>12$), medium ($8<\text{score}\leq13$), and low ($\text{score}\leq8$) were scored by the program coordinator and provided to the model as reference examples. These examples served as anchors, allowing iterative refinement of the prompt to adjust scoring strictness or leniency. This few-shot approach significantly improved rubric alignment with human raters compared to zero-shot approaches \citep{tian2024examining, metzler2024computer, rao2026autorubric}. The calibration process was iterative. Example outputs were reviewed, the prompt was adjusted to address scoring deviations, and the examples were re-run until the model's outputs matched the ground truth scores for the reference SoPs.

\subsection{Human Review and Candidate Pool Selection}

Following LLM-based scoring, the full set of approximately 1,200 scored and annotated outputs would be passed to the program coordinator for review alongside each applicant's SoP. The LLM outputs were designed to replicate the role previously played by distributed human graders, providing the coordinator with scored and rationale-annotated submissions for the full applicant pool. The human review step was considered essential, as the outputs served as a structured first pass to guide review order. Final candidate pool selection remained entirely with the program coordinator, applying the standard selection criteria used in prior program cycles.

\section{Preliminary Observations}

\subsection{Processing and Review Workflow}

\begin{figure}[h]
    \centering
    \includegraphics[width=\textwidth]{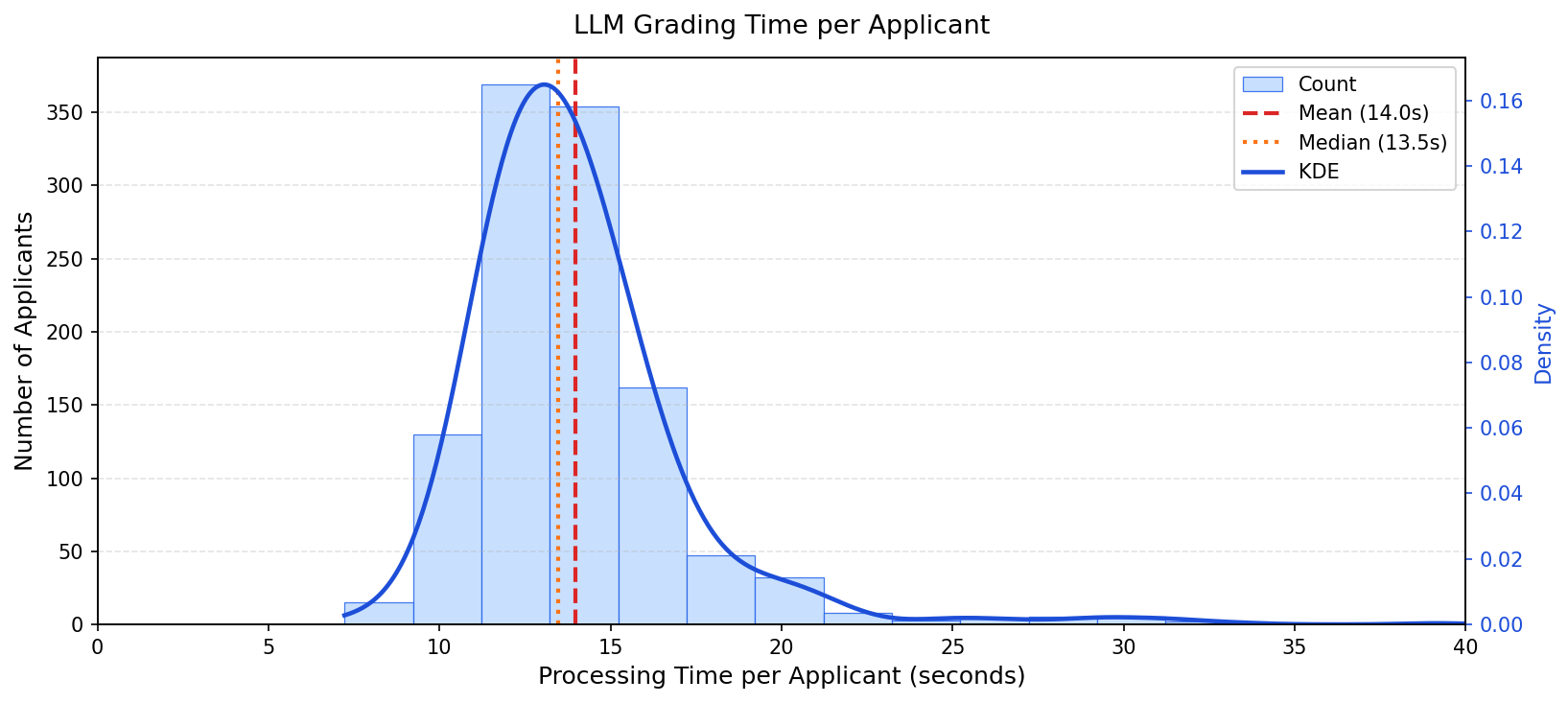}
    \caption{LLM Grading time per applicant (GPT-5.2)}
    \label{fig:processing_time}
\end{figure}

The approximately 1,200 eligible SoPs were processed via the OpenAI API using a custom script\footnote{Source code available at \url{https://github.com/Salazar-Prime/llm-sop-rubric-evaluator}}, with each evaluation against GPT-5.2 averaging approximately 14 seconds (per-SoP timing varied with SoP length, which ranged from 500 to 2,000 words). The full batch was processed in approximately 4.6 hours of compute time and run overnight, with results stored in a spreadsheet (Figure \ref{fig:processing_time}). The total cost of API calls for GPT-5.2 was \$25 for all 1,200 SoPs. Each record contained the applicant's scores across the six sub-categories, detailed rationale text with direct citations from the SoP, and a total score. The program coordinator then reviewed applicants in order of total score, with particular attention to the rationale and cited evidence for each evaluation. High-potential applicants were flagged on the program's selection portal. This workflow enabled a single coordinator to complete the review in approximately 4 hours, compared to the multi-week effort previously required to recruit and coordinate a team of human graders and aggregate their results. Addition of cited evaluation rationale, aided the coordinator to quickly scan the cited text and confirm that the score was justified, rather than re-reading the entire SoP.

\subsection{Model Version Differences}

During development, multiple OpenAI model versions were tested, including GPT-4o, GPT-5-mini, and GPT-5.2. A notable finding was that GPT-5.2 demonstrated substantially stronger adherence to the structured rubric compared to GPT-4o, particularly in its treatment of lower-scoring submissions. GPT-4o tended to produce more lenient or inconsistent scores for weaker SoPs, while GPT-5.2 applied the rubric criteria more uniformly. GPT-5-mini was also tested but was found to be inadequate for this evaluation task, producing outputs that lacked the necessary depth of analysis. This observation aligns with the broader literature showing that model version and capability significantly affect scoring reliability \citep{tang2024harnessing, mathew2026llms}.

\subsection{Inter-model Score Divergence}

\begin{figure}[h]
    \centering
    \includegraphics[width=\textwidth]{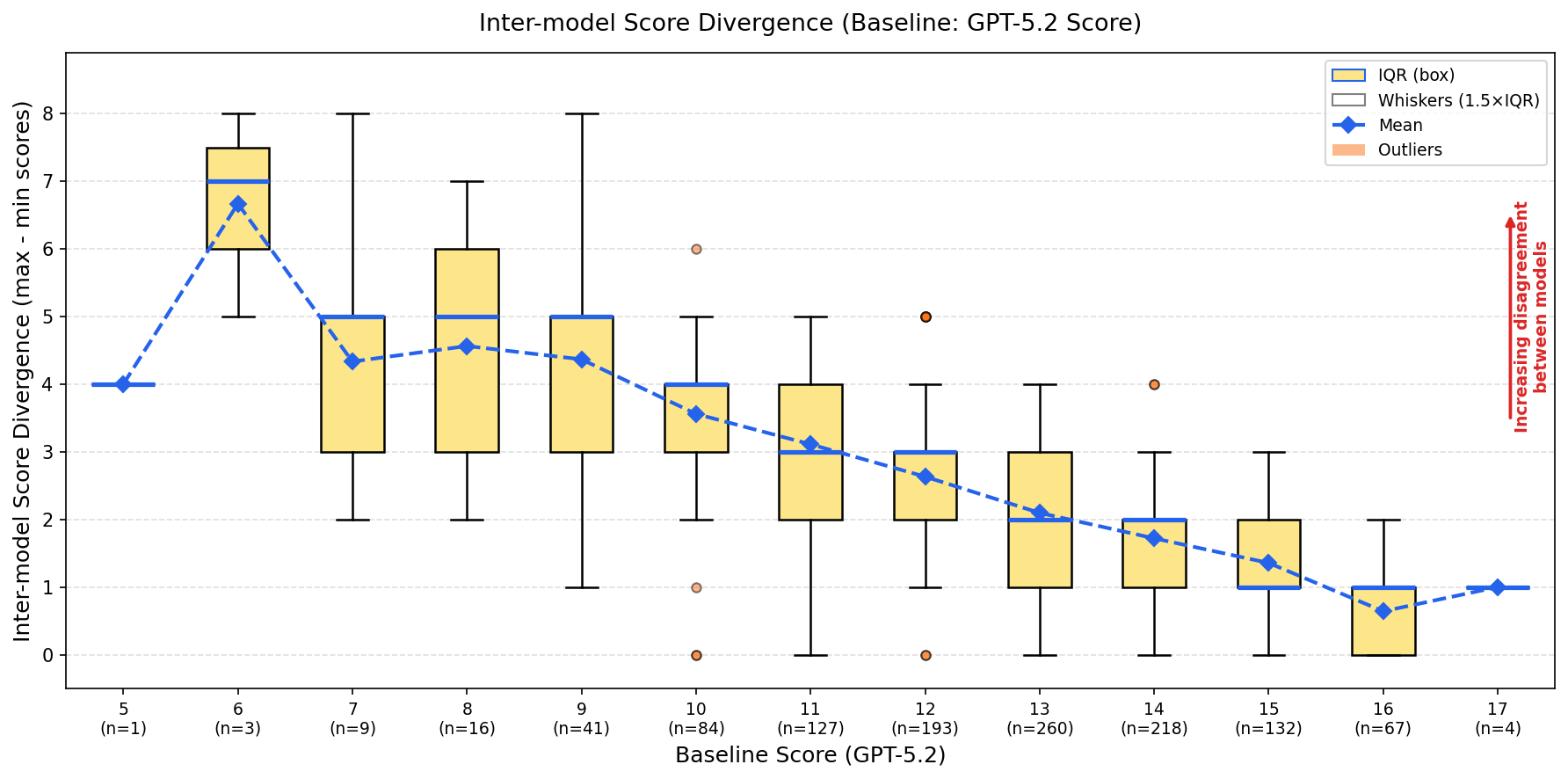}
    \caption{Inter-model Score Divergence with GPT-5.2 Score as baseline. Box plots show the distribution of disagreement (max - min scores) among GPT-5.2, GPT-5-mini, and GPT-4o for each applicant, grouped by their GPT-5.2 baseline score. Greater model disagreement is seen in lower scoring applicants. Sample sizes (n) are shown below each baseline score.}
    \label{fig:score_spread}
\end{figure}

To assess inter-model consistency, we compared the total scores produced by GPT-4o, GPT-5-mini, and GPT-5.2 across the 1,200 evaluated SoPs (Figure \ref{fig:score_spread}). Inter-model Score Divergence, defined as the difference between the maximum and minimum score assigned by any of the three models for a given applicant, was used as a measure of disagreement. Inter-model divergence was substantially lower for high-scoring submissions than for low-scoring ones. For applicants receiving a GPT-5.2 total score of 13 or above, the mean score spread across models was approximately 2 points or less, indicating strong consensus. By contrast, applicants scoring below 10 exhibited mean spreads of 4 points or higher, with considerable variance. This suggests that the rubric and prompt provided sufficient guidance for the models to converge on strong submissions, where indicators of passion, initiative, and resilience are clearly and consistently expressed, but that weaker submissions, where such qualities are absent or ambiguously stated, leave more room for divergent model interpretations. These findings are consistent with broader observations in the automated essay scoring literature that LLM scoring reliability tends to decrease at the lower end of the performance distribution \citep{mathew2026llms, mansour2024can}.

\section{Practical Impact and Limitations}

The most immediate practical impact of the tool was the reduction in coordination overhead. In prior SURF cycles, the review process required recruiting and training a team of human graders, distributing applications, monitoring progress, collecting and reconciling scores, and resolving discrepancies. With the LLM-based approach, the evaluation of all 1,200 SoPs was completed overnight (approximately 4.6 hours of GPT-5.2 compute time), and a single program coordinator spent approximately 4 hours reviewing the sorted results and flagging high-potential applicants.

A few practical lessons emerged from this deployment that may be useful for other practitioners building similar tools. First, defining adjacent score levels in terms of objectively observable behaviors, rather than qualitative impressions, produced more consistent model scoring and made human verification substantially easier. Second, requiring the model to support each score with a direct citation from the submission was the single most valuable design choice for human review; it converted score-checking from a re-reading task into a quick verification of cited evidence. Third, few-shot calibration with a small number of human-scored examples at each performance level was sufficient to align model outputs with program expectations, but the calibration set itself needed to be revisited when the underlying model version changed. Fourth, iterating on the rubric in response to observed model outputs, rather than treating the rubric as fixed up front, was essential to reaching reliable performance. A more systematic study of how changes in rubric wording affect scoring behavior is left to future work.

This work has several important limitations. First, the comparison between model and human scores was informal rather than systematic; a formal inter-rater reliability analysis (for example, using Quadratic Weighted Kappa) was not conducted. Second, the tool was used to flag high-potential applicants on the selection portal, but final program selection involved additional criteria beyond the SoP evaluation, so the tool's impact on ultimate selection outcomes cannot be isolated. Third, while the prompt was designed to reduce bias by applying the same rubric uniformly, LLMs may carry biases from their training data that affect how they evaluate different writing styles, disciplinary contexts, or cultural communication norms that can be harder to isolate \citep{kasneci2023chatgpt}.

\section{Ethical Considerations}

Deploying LLMs in a high-stakes evaluation context such as application review introduces several ethical considerations that practitioners should weigh before adopting a similar workflow.

\paragraph{Student data privacy and model training.} Statements of Purpose contain personal narratives, family circumstances, and details of academic and professional history that students reasonably expect to remain confidential. Before using any LLM provider for this kind of evaluation, the program staff must understand the provider's data retention and model training policies and ensure that applicant submissions are not used to train future models. In practice, this means using API tiers that contractually exclude submitted content from training corpora (such as the paid OpenAI API calls rather than web based chat interfaces), and confirming that data is not retained beyond the period required for review process. Where possible, identifying information (names, contact details, institution names) should be removed before submission to the model. Notably, however, Statements of Purpose contain identifying details that are difficult to fully redact through automated means, such as descriptions of unique research projects or named mentors, so privacy-preserving redaction remains an open challenge for this use case.

\paragraph{Human-in-the-loop is non-negotiable.} The workflow described here uses the LLM as a triage and ranking aid, not as a final decision-maker. All flagging and selection decisions remain with the program coordinator, who reviews the model's score together with its cited evidence before any applicant is advanced. We consider this human review essential rather than optional: without it, model biases or scoring errors propagate directly into selection outcomes, and applicants lose meaningful recourse.

\paragraph{Fairness and the comparison to human review.} It is important to note that human reviewers are themselves not perfectly consistent. A growing body of evidence shows that human grading varies with factors unrelated to applicant merit, including time of day, reviewer fatigue, and external conditions \citep{vicario2025timing, gurung2022considerate, mehta2025morning}. The relevant question is therefore not whether the LLM is perfect, but whether the combination of LLM scoring with human verification is at least as fair and consistent as the prior human-only process, while remaining accountable. Even so, LLMs may carry biases from their training data that affect how they evaluate different writing styles, disciplinary contexts, or cultural communication norms \citep{kasneci2023chatgpt}. Requiring the model to ground each score in a direct citation from the submission, as described in Section~\ref{sec:rubric-design}, partially mitigates this risk by making bias-driven scoring decisions easier to detect during human review, but does not eliminate it. A formal fairness audit across applicant subgroups remains an important direction for future work.

\section{Future Work}

Several directions for future work are planned. First, conduct a formal validation study comparing LLM scores to a panel of human reviewers using established inter-rater reliability metrics. Second, investigate whether the model's weaker performance on lower-scoring submissions can be improved through additional calibration examples at the low end of the scoring distribution. Third, explore the development of a web-based interface (under-development) that would allow program staff without programming expertise to use the tool with custom rubrics. The same rubric-driven workflow generalizes to other writing artifacts common in undergraduate research programming, such as CVs, research abstracts, reflections, and graded course assignments. Each of these contexts carries its own scoring nuances that can be encoded directly in the rubric: for example, in abstract grading, technical jargon would not be penalized to the same degree as it would be in an SoP. We anticipate that program staff will be able to adapt the rubric to their individual program requirements.

\section{Conclusion}

This work-in-progress paper describes the development and initial deployment of an LLM-based tool to support high-volume application review for an undergraduate research program. By combining a structured rubric assessing Passion, Clarity of Purpose, and Resilience, carefully engineered prompts with citation requirements, few-shot calibration, and a human-in-the-loop verification workflow, the tool replicated the role of distributed human graders, providing the program coordinator with scored and rationale-annotated outputs for the full applicant pool. The coordinator then applied standard program criteria to this output to produce a shortlist of strong candidates, the same process used in prior cycles, but with the LLM replacing the coordination overhead of recruiting, training, and reconciling scores from a team of human graders. This enabled a single program coordinator to complete the full review and identify a candidate pool in a fraction of the time previously required. Preliminary observations suggest that current LLMs can serve as effective triage tools for application review when provided with explicit rubrics and grounded evaluation requirements, though their reliability varies across model versions and performance levels of the submissions being evaluated. Importantly, the LLM outputs were not used to directly shortlist applicants at any stage; human judgment remained central to all selection decisions. This approach has broad applicability to other high-volume evaluation contexts in program processes and engineering education.

\bibliographystyle{unsrtnat}
\bibliography{references}

\appendix
\newpage
\section*{Appendix A: Evaluation Rubric}
\addcontentsline{toc}{section}{Appendix A: Evaluation Rubric}

The following rubric was part of the evaluation prompt. It defines three main categories (Passion, Clarity of Purpose, and Resilience), each with two sub-categories scored on a 0-3 scale, for a maximum total score of 18. Each score level includes a behavioral descriptor that served as an anchor for the model's evaluation.

\paragraph{Category 1: Passion.}
\leavevmode\\
\\
\textit{Motivation for Scientific Research - Why are they interested in STEM/research?}
\begin{itemize}
\item \textbf{0:} The candidate gives no clear indication of what sparked their interest in STEM or why they continue to pursue it.
\item \textbf{1:} The candidate mentions either their interest in STEM or their motivation to continue in STEM, but not both.
\item \textbf{2:} The candidate mentions both their interest in STEM and their motivation to continue in STEM.
\item \textbf{3:} The candidate discusses both their interest in STEM and their motivation to continue in STEM with clear enthusiasm and personal experiences that highlight their passion.
\end{itemize}

\textit{Initiative - Seeks out and pursues opportunities to engage with their chosen field.}
\begin{itemize}
\item \textbf{0:} Only seeks scientific interests through coursework.
\item \textbf{1:} Has taken part in at least one other opportunity outside of mandatory coursework.
\item \textbf{2:} Actively looks for other opportunities, taking advantage of networking or applying for opportunities through competitive application processes.
\item \textbf{3:} Actively seeks opportunities outside of their home institution to deepen understanding and engagement in STEM.
\end{itemize}

\paragraph{Category 2: Clarity of Purpose.}
\leavevmode\\ \\
\textit{Program Expectations and Benefits - Demonstrates a realistic comprehension of program expectations and benefits.}
\begin{itemize}
\item \textbf{0:} Does not acknowledge why chose SURF program or Purdue.
\item \textbf{1:} Mentions SURF and/or Purdue but gives few details or drastically misunderstands the scope.
\item \textbf{2:} Describes why they are interested in SURF, specifically mentioning professional development opportunities OR mentions a department or faculty at Purdue they are interested in.
\item \textbf{3:} Describes why they are interested in SURF, specifically mentioning professional development opportunities AND mentions a department or lists specific faculty at Purdue.
\end{itemize}

\textit{Alignment with Future Endeavors and Career Direction - Provides a clear connection between the program and their future goals.}
\begin{itemize}
\item \textbf{0:} No future goals are mentioned.
\item \textbf{1:} May or may not specifically mention SURF/Purdue, but briefly fits into non-graduate school career plans.
\item \textbf{2:} Describes how SURF will help them prepare for their interest in graduate school.
\item \textbf{3:} Describes in detail the PhD or MS degree they want to pursue in graduate school with mention of a particular field.
\end{itemize}

\paragraph{Category 3: Resilience.}
\leavevmode\\ \\
\textit{Reflection on Learning through Experience - Showcases adaptability and tenacity by reflecting on personal, academic, or research challenges.}
\begin{itemize}
\item \textbf{0:} Does not mention a challenge.
\item \textbf{1:} Mentions the challenge without reflection on how it helped in self-growth.
\item \textbf{2:} Provides an example of a personal challenge they faced and reflects on it.
\item \textbf{3:} Provides an example of a prolonged personal challenge due to societal issues and demonstrates perseverance through it.
\end{itemize}

\textit{Problem Solving - Demonstrates problem-solving skills by addressing challenges and overcoming setbacks.}
\begin{itemize}
\item \textbf{0:} Does not mention a challenge or demonstrates an inability to cope with challenges.
\item \textbf{1:} Mentions the challenge without mentioning how the solution was reached using their own abilities.
\item \textbf{2:} Mentions a personal challenge and lays out their strategy, possibly by breaking it down into more manageable tasks.
\item \textbf{3:} Provides an example of a personal challenge with a detailed, laid-out step-by-step plan for addressing it. Additionally, actively seeks outside resources and possibly forges new relationships to tackle the issue.
\end{itemize}

\section*{Appendix B: Evaluation Prompt}
\addcontentsline{toc}{section}{Appendix B: Evaluation Prompt}

The following is the complete prompt sent to the language model for each SoP evaluation. The prompt was sent as the user message in the OpenAI chat completions API call. The rubric (Appendix A), serialized as JSON, was embedded directly within this prompt following the instruction text. Few-shot calibration examples (not shown) were provided as prior conversation turns.

\begin{lstlisting}
Evaluate the following Statement of Purpose (SoP) by using the specified rubric.
Your output should be in JSON schema, containing a score for each category
and a rationale supported by citations from the SoP. A description of the
score is provided in the prompt that will help you determine the score. The
rationale should have both positive and negative aspects that helps justify
the score. If the candidate is on the borderline, then fractional score is
allowed.

[The full rubric JSON is appended - see Appendix A]

[Applicant SoP]
\end{lstlisting}

\section*{Appendix C: Prior (2024) Evaluation Rubric}
\addcontentsline{toc}{section}{Appendix C: Prior (2024) Evaluation Rubric}

The following rubric was used in earlier SURF cycles prior to the 2026 redesign described in this paper. It is reproduced here verbatim from the 2024 program documentation to illustrate the shift from subjective to objectively observable score-level descriptors discussed in Section~\ref{sec:rubric-design}. The 2024 rubric was authored for human reviewers and relied on qualitative prompts (for example, ``contagious enthusiasm,'' ``exceptional problem-solving abilities''), which proved harder for the LLM to apply across submissions.

\paragraph{Category 1: Passion.} Passion is a key factor in evaluating candidates, encompassing both the motivation to pursue STEM and demonstrated initiative. We seek individuals who not only articulate what sparked their interest in the STEM field but also showcase a deep connection between their background and the pursuit of their chosen subject. The ideal candidate will display genuine enthusiasm for scientific research, actively engaging in STEM-related activities with a contagious desire to explore and contribute to scientific knowledge.
\leavevmode\\ \\
\textit{Motivation for Scientific Research - Why are they interested in STEM/research?}
\begin{itemize}
\item \textbf{0 - Poor:} The candidate gives no clear indication of what sparked their interest in STEM or why they continue to pursue it.
\item \textbf{1 - Developing:} Sporadic or superficial interest in science. They may mention why or how they got into STEM, but provides no indication of how they're fueled to continue on in research.
\item \textbf{2 - Proficient:} Describes how they became interested in STEM and discusses their current motivation to pursue their research area.
\item \textbf{3 - Excellent:} Thorough, passionate description of entering STEM and current motivation. Displays contagious enthusiasm for their field, and are excited to contribute to it.
\end{itemize}

\textit{Initiative - Seeks out and pursues opportunities to engage with their chosen field.}
\begin{itemize}
\item \textbf{0 - Poor:} Takes no initiative in pursuing scientific interests beyond mandatory coursework, may even pass up clear opportunities to engage (e.g., advanced students at an R1 research institution not taking advantage of the research on their own campus).
\item \textbf{1 - Developing:} Engages with opportunities easily available to them, but doesn't exert much effort to find advanced opportunities (e.g., only participated in course-based research or minimal engagement with readily available faculty research on their campus).
\item \textbf{2 - Proficient:} Takes initiative in engaging in STEM extracurriculars. Demonstrates proactivity within the constraints of their academic setting (e.g., worked with a lab on their campus if available; joined clubs, attends seminars).
\item \textbf{3 - Excellent:} Actively seeks out opportunities to deepen understanding and involvement in STEM, showcasing resourcefulness and determination (e.g., has numerous or ongoing research experiences at their home institution if available, and/or has taken part in external research programs, possibly applied for grants or fellowships; at a small liberal arts school this may look like a passion project or a self-driven one-on-one project with a faculty member).
\end{itemize}

\paragraph{Clarity of Purpose.} Clarity of purpose is crucial in assessing candidates' understanding of the Summer Undergraduate Research Fellowship program. Successful applicants will demonstrate a realistic comprehension of program expectations and benefits, showcasing an awareness of specific components and the significance of potential benefits. Furthermore, candidates should articulate a clear connection between the program and their future endeavors, providing detailed insights into how the program aligns with their career goals and academic aspirations.
\leavevmode\\ \\
\textit{Program Expectations and Benefits.}
\begin{itemize}
\item \textbf{0 - Poor:} Does not acknowledge why this program or this university. Provides only a generic personal statement that could be used to apply to any program.
\item \textbf{1 - Developing:} Vague or unrealistic expectations of the program. Mentions SURF and/or Purdue, but either gives few details about why, or drastically misunderstands scope of program (e.g., expects to publish a paper with a faculty they've never met by the end of the summer).
\item \textbf{2 - Proficient:} Describes why they are interested in SURF, specifically mentioning professional development opportunities OR mentions a department or faculty at Purdue they are interested in working with and why.
\item \textbf{3 - Excellent:} Describes why they are interested in SURF, specifically mentioning professional development opportunities AND mentions a department or lists specific faculty at Purdue they are interested in working with and why.
\end{itemize}

\textit{Alignment with Future Endeavors and Career Direction.}
\begin{itemize}
\item \textbf{0 - Poor:} Does not describe future goals or does not mention how a summer research program contributes to achieving them.
\item \textbf{1 - Developing:} May or may not specifically mention SURF/Purdue, but briefly describes how a summer research experience fits into vague or undeveloped career plans.
\item \textbf{2 - Proficient:} Describes how SURF will help them prepare for their future, and expresses interest in graduate school in the personal statement.
\item \textbf{3 - Excellent:} Describes what graduate degree they want to pursue and recognizes that SURF can either help them hone their research skills in that field or help them narrow in on a specific research area of interest.
\end{itemize}

\paragraph{Resilience.} Resilience is a crucial trait for research, given its iterative nature. Candidates are expected to reflect on personal, academic, or research challenges, showcasing adaptability and tenacity. Successful applicants demonstrate a strong growth mindset and an ability to both navigate and grow from encountering setbacks.
\leavevmode\\ \\
\textit{Reflection on Learning through Experience.}
\begin{itemize}
\item \textbf{0 - Poor:} Actively avoids difficult situations, quits when things get hard.
\item \textbf{1 - Developing:} May mention something that would have been difficult to the average person but does not reflect on their experiences or engage with their environment to learn and grow from it.
\item \textbf{2 - Proficient:} Reflects on hardships they have faced and describes how they have personally grown from them.
\item \textbf{3 - Excellent:} Describes an exceptionally difficult situation they have faced and reflects on what it has taught them AND/OR specifically mentions that facing difficulty provides an opportunity to grow, and eagerly approaches challenges with this mindset.
\end{itemize}

\textit{Problem Solving.}
\begin{itemize}
\item \textbf{0 - Poor:} Demonstrates an inability to cope with challenges; easily discouraged by setbacks. May avoid addressing difficulties.
\item \textbf{1 - Developing:} Acknowledges facing problems and attempts to address them, although may do so poorly or inefficiently.
\item \textbf{2 - Proficient:} Identifies and articulates a challenge they faced. Strategically works to address it, possibly by breaking it down into more manageable tasks or seeking guidance from their immediate community (e.g., lab mentor, PI) and effectively utilizes available resources.
\item \textbf{3 - Excellent:} Demonstrates exceptional problem-solving abilities by systematically breaking down complex problems into smaller tasks and developing a strategic plan for resolution. Not only takes advantage of immediate community and resources, but actively seeks outside resources, possibly forging new relationships or methods to address the problem.
\end{itemize}

\end{document}